\documentclass[letterpaper,twocolumn,fleqn]{article} 

\usepackage{ist}
\usepackage{times}
\usepackage{soul}
\usepackage{url}
\usepackage[utf8]{inputenc}
\usepackage[small]{caption}
\usepackage{graphicx}
\usepackage{amsmath}
\usepackage{amsthm}
\usepackage{booktabs}
\usepackage{algorithm}
\usepackage{algorithmic}
\usepackage{epsfig}
\usepackage{subcaption}
\usepackage{amssymb}
\usepackage{color}
\usepackage{mathtools}
\usepackage{enumitem}

\title{Conditional Synthetic Food Image Generation}

\author{Wenjin Fu\textsuperscript{1}, Yue Han\textsuperscript{2}, Jiangpeng He\textsuperscript{2}, Sriram Baireddy\textsuperscript{2}, Mridul Gupta\textsuperscript{2}, Fengqing Zhu\textsuperscript{2}\newline\newline 
\textsuperscript{1} School of Computer Science and Engineering, The Ohio State University, Columbus, OH, United States \newline
\textsuperscript{2}Elmore Family School of Electrical and Computer Engineering, Purdue University, West Lafayette, Indiana, United States \newline}

\date{Feb 6 2023} 

\begin{document} 

\maketitle 

\thispagestyle{empty} 


\begin{abstract}
Generative Adversarial Networks (GAN) have been widely investigated for image synthesis based on their powerful representation learning ability. In this work, we explore the StyleGAN and its application of synthetic food image generation. Despite the impressive performance of GAN for natural image generation, food images suffer from high intra-class diversity and inter-class similarity, resulting in overfitting and visual artifacts for synthetic images.
Therefore, we aim to explore the capability and improve the performance of GAN methods for food image generation. Specifically, we first choose StyleGAN3 as the baseline method to generate synthetic food images and analyze the performance.
Then, we identify two issues that can cause performance degradation on food images during the training phase: (1) inter-class feature entanglement during multi-food classes training and (2) loss of high-resolution detail during image downsampling. To address both issues, we propose to train one food category at a time to avoid feature entanglement and leverage image patches cropped from high-resolution datasets to retain fine details. We evaluate our method on the Food-101 dataset and show improved quality of generated synthetic food images compared with the baseline. Finally, we demonstrate the great potential of improving the performance of downstream tasks, such as food image classification by including high-quality synthetic training samples in the data augmentation.
\end{abstract}

\section{Introduction}
Healthy diet is one of the key factors for human  wellness and disease prevention. There is a growing trend for people to track their dietary intake to adhere to or maintain a healthy diet. Traditional dietary assessment methods~\cite{traditional_record, traditional_record2} rely on manual self-reporting, which can be tedious and time-consuming. Image-based dietary assessment \cite{shao2021_ibdasystem, bib5} aims to develop automated methods to analyze consumed food types\cite{mao2020visual, mao2021_nutri_hierarchy}, portion size~\cite{shao2021towards, he2020multitask, he2021end} directly from captured eating occasion images. One of the major challenges of image-based dietary assessment is the lack of enough food images in existing datasets~\cite{shao2021_nutri_database, han2021improving} to train a robust deep learning model for food analysis. For example, the food recognition performance on less commonly seen food categories could drop significantly~\cite{zhang2021deep, zhang2022learning} due to the few available training data. Many efforts have been made to solve the problem of lacking enough food images, such as for long-tailed classification~\cite{he2022long} to address severe class-imbalance issue, continual learning~\cite{ILIO, food_ocil, raghavan2023online} to learn from new data incrementally, and other food analysis scenarios~\cite{jiang2020few, pan2022_madima} that focus on real world food data distribution. 

Generative network is widely applied as an effective data augmentation method to help address the issue of insufficient training data. Over the years, generative networks have been revolutionized from a basic autoencoder for reconstructing input data to a learning feature representation for creating non-existent objects. In recent years, the paradigms of the state-of-the-art generative models focus on three structures: Variational Autoencoders (VAEs) \cite{bib8} (VDVAE \cite{bib9} offers high image diversity), Diffusion models \cite{bib10} (DDPM2 \cite{bib11} offers advanced image quality and variety, but low sampling speed), and Generative Adversarial Networks (GANs) \cite{bib12} (StyleGAN \cite{bib13} offers good image quality and sampling speed). In general, GANs have been demonstrated to generate high-fidelity synthetic images efficiently.



Food image synthesis using GAN has been widely investigated such as CookGAN \cite{bib14}, built on a cycle-consistent network \cite{bib15}, RamenGAN \cite{bib16}, built on a standard conditional network \cite{bib17}, and multi-ingredients pizza generator \cite{bib18}, built on StyleGAN2 \cite{bib19} have shown a decent performance on food image generation. However, the food images generated by these methods either do not provide sufficient details or contain many artifacts. Among existing GAN methods, StyleGAN3 \cite{bib20} shows an ability to generate highly realistic images. In this work, we explore StyleGAN3 with its capability of generating food images corresponding to their labels. 

Despite several improvements had been made in StyleGAN3, we discovered two issues could be addressed when generating synthetic food images: (1) inter-class feature entanglement (the generated image for a specific class contains features from other image classes) and (2) loss of high-resolution details during data normalization (e.g., image size rescaling and downsampling). Then, we propose two training strategies to address these issues, including single-class training to avoid features being correlated between different classes, and image-patches training on any-resolution data to avoid image normalization. We evaluate our proposed method on the Food-101 dataset ~\cite{food101} with the Frechet Inception Distance Metric (FID) \cite{bib22} and a subjective survey to demonstrate the effectiveness in improving the visual resolution and fidelity of our generated food images. Finally, we use our synthetic food images as additional training images for training a food image classifier to explore the impact of data augmentation using synthetic images. 



\section{Preliminaries} 
The main idea of Generative Adversarial Networks\cite{bib11} is to train a generator network (G) that maps the noise vectors to real training data distribution to create realistic image instances. Meanwhile, the discriminator model (D) attempts to distinguish the real data from generated samples via an estimated probability. The two networks are optimized simultaneously during training.

Following the idea of GAN, StyleGAN3 \cite{bib20} also uses a generator to generate synthetic images and a discriminator to distinguish real from synthetic samples. However, the generator network is made more complex. Instead of directly feeding the noise vector to the generator, StyleGAN3 goes through a mapping network to reduce correlation among different features during training. With different combinations of style (feature) information learned from the network, StyleGAN3 has a synthesis framework, which composes 14 layers to collect and generate coarse and fine styles sequentially to generate high-quality synthetic images. The improvements of StyleGAN3 also include solving the feature adhesion under coarse layers and making the generation process invariant to image translation and rotation.



In order to generate synthetic food images corresponding to their class label, we investigate conditional image generation where the image class labels are supervised during training. The StyleGAN3 network controls the generation of image class from two basic parts: the mapping network in the generator and the discriminator network. The mapping network conditions the latent code with a one-hot label vector which defines a set of specific characteristics from a certain class for the generator to study, while the discriminator is trained to classify real and generated data conditioned on their class labels. Therefore, with a feature vector to control the image's underlying content spatial structure, the generator can generate synthetic images for specific classes. 




\section{Proposed Methods}

The proposed methods aim to improve the training of StyleGAN3 for generating realistic synthetic food images. The first method involves training with a single-class food dataset to avoid feature entanglement, while the second method involves training with any-resolution data to capture fine-grain details in high-resolution images. These approaches have the potential to address specific challenges in training and enhance the performance of StyleGAN3.

\textbf{Training StyleGAN3 with a single-class at a time.} 
According to the results of training StyleGAN3 on low-resolution multi-class food datasets in the Experiment section, we find that even though the conditional StyleGAN3 model is trained to stabilize and converged based on the FID metric evaluation on generated synthetic images, the results of generated synthetic food images still look unnatural and the reason of artificial-looking and distorted synthetic images are caused by inter-class feature entanglement (e.g., Figure 
 \ref{fig:inter-class-issue} shows that synthetic hamburger images include features from spring roll). Either features in different classes are not well-distributed in the mapping network, or the discriminator could not classify the real and synthetic images into their perspective classes due to complex and similar features learned in different classes. To avoid features being correlated and affecting each other, we trained StyleGAN3 with a single-class food dataset one at a time to avoid feature entanglement.

\textbf{Training StyleGAN3 with any-resolution data.} After analyzing the results from StyleGAN3, we found that this baseline method has a few drawbacks. To train a network, the input images have to be fixed at certain resolutions, such as $256 \times 256$, $512 \times 512$, or $1024 \times 1024$.
This requirement could lead to image warping and loss of image details when downsampling the input images from high-resolution to the required low-resolutions. To avoid losing fine-grain details in high-resolution images during downsampling, we adopt the method from Anyres GAN \cite{bib21} to project and capture previously discarded high-resolution image details.

\begin{figure}[!hb]
  \centering
  \includegraphics[width=1\columnwidth]
  {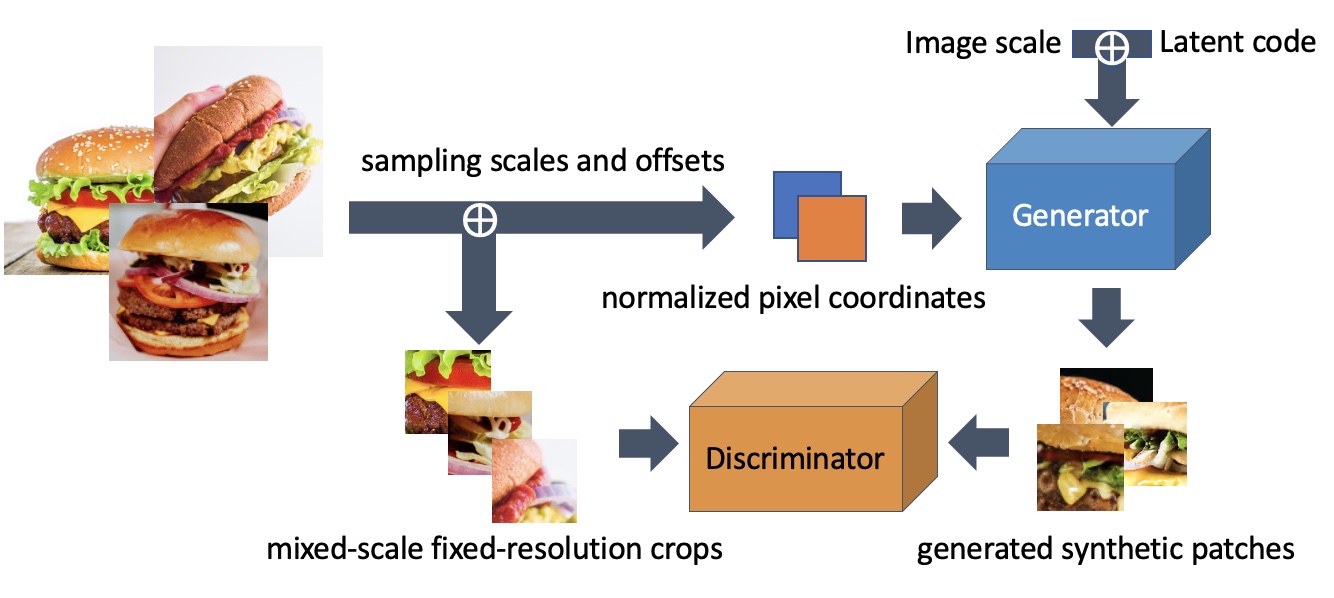}
  \caption{\textbf{Training Architecture of Anyres GAN.}}
  \label{fig:anyres}
\end{figure}

More specifically, Anyres GAN \cite{bib21} includes two stages of generator training: global fixed-resolution pretraining and mixed-resolution patch-based training. In the first stage, the network follows the standard training procedure of StyleGAN3, which is trained on $256 \times 256$ resolution of single food class images (e.g., images from hamburger class) to capture the global structure of the given training images. This model is then used as a teacher model during the second stage of training. To better learn the fine-grain details for synthetic images, we randomly crop square-shaped patches from the same class of high-resolution images with any resolutions in the second stage of training. These images include synthetic food images generated by our pretrained StyleGAN3 and high-resolution images scrapped from Google at various random scales and locations. The generator takes three inputs: normalized pixel coordinates of our sampled patches, the original image resolution which the patches are created from, and the latent code $z$ representing the underlying features of the original image for the generator to produce synthetic square patches.

The discriminator compares the synthetic patches with real patches to help the generator for obtaining fine details in generated image patches. The generated patches are then adjusted to match the teacher's global fixed-resolution output after proper downsampling and alignment. In the end, the patch features are projected into global fixed low-resolution images to obtain fine details in those high-resolution images. The training architecture of the Anyres GAN is illustrated in Figure \ref{fig:anyres}.


\section{Experiments}

In this section, we evaluate our proposed method by conducting different experiments on low-resolution multi-class datasets, low-resolution single-class datasets, and any-resolution single-class food datasets. In addition, we perform both objective and subjective tests to show the perceptual visual realism of our generated food images. Finally, we demonstrate the effectiveness of using synthetic data as data augmentation to improve the performance of food image classification.


\subsection{Datasets}


\textbf{Low-resolution Multi-class and Single-class Food Dataset} 
We construct a dataset with ten random food classes selected from the Food-101 dataset ~\cite{food101} for evaluating the baseline conditional StyleGAN3. These ten food classes include cannoli, cupcake, donut, hamburger, pancake, strawberry, shortcake, pizza, spring roll, panna cotta, and waffle images. 
Each class contains $1,000$ images, and each image has a maximum resolution of $512$ pixels and a minimum resolution of $384$ pixels. We pre-process the images to a dimension of $256 \times 256$ to meet the input image resolution requirement of StyleGAN3 and refer to this dataset as the low-resolution food dataset (LR). In contrast, the low-resolution single-class Food Dataset consists of only hamburger food images from Food-101 downsampled to $256 \times 256$.

\textbf{Any-resolution Dataset for Anyres Training.} 
We use $600$ high-resolution hamburger images scraped from Google as part of our selected dataset for training the second stage of any-resolution GAN. In this dataset, the minimum side length is $512$ pixels, the maximum side length is $5,472$ pixels, the mean side length is $1,250.06$ pixels, and the median side length is $1,000$ pixels. We combine the images from the Food-101 dataset and the high-resolution hamburger images from Google to form the selected Any-resolution dataset where all images have a resolution greater than $256$. During image-patches training, image patches with $256$ resolution are cropped from both Any-resolution and LR datasets.




\subsection{Evaluation Metrics}
The Frechet Inception Distance (FID) \cite{bib22} is a commonly used metric to evaluate the similarity between the distribution of real and synthetic images. The lower the FID scores, the more realistic of generated images are. The FID metric is been shown to be computationally efficient and consistent with human assessment of synthetic image discrimination \cite{bib23}. In our experiment, we compute the FID for every five training epochs based on the saved model and sample results. In addition, we also conducted a subjective study to qualitatively evaluate our model, where we ask 82 adult participants to assess the perceptual realism of our synthetically generated food images.


\subsection{Results on Low-resolution Multi-class Food Datasets}
\label{subsec:low-multi}
\begin{figure}[!h]
  \centering
  \includegraphics[width=0.7\columnwidth]
  {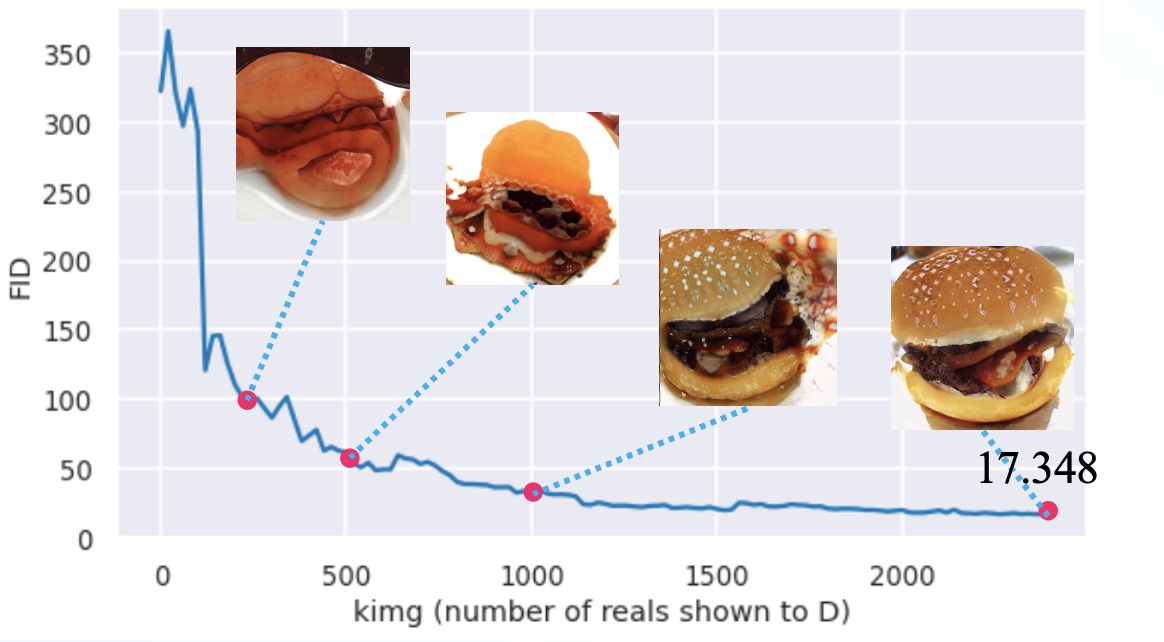}
  \caption{\textbf{Evaluation of Synthetic Multi-Class Food Images on conditional StyleGAN3}}
  \label{fig:multi-food-training}
\end{figure}

During conditional StyleGAN3 training, we calculate the FID score to evaluate the network's performance and report the lowest FID metrics. As shown in Figure \ref{fig:multi-food-training}, we select the synthetic hamburger images as a representative class to show that the FID metric effectively evaluates the efficiency of the StyleGAN3 network and the visual quality of the generated image. We record the FID score for every $20$ iteration of training until the network converged at the score of $17.348$. 

\begin{figure}[!h]
  \centering
  \includegraphics[width=0.9\columnwidth]
  {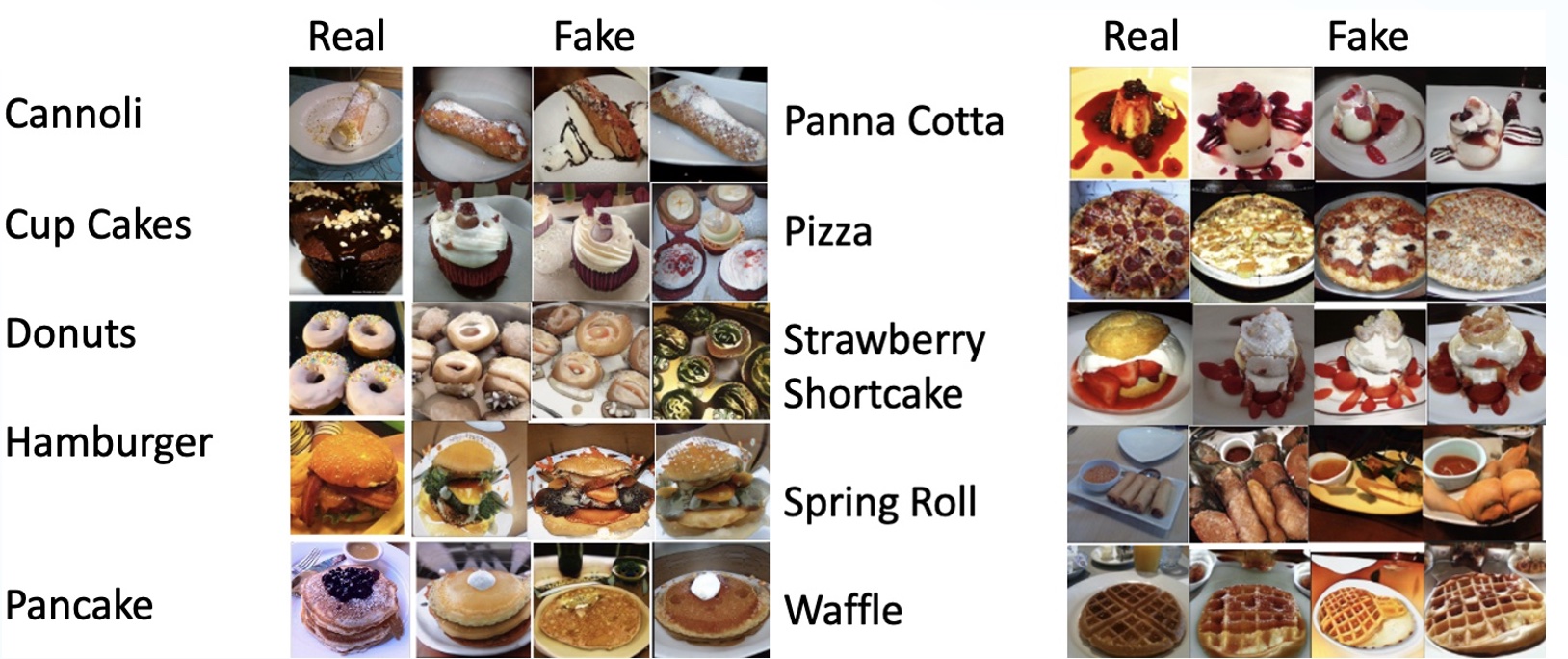}
  \caption{\textbf{Example of conditional synthetic images results from global fixed resolution}}
  \label{fig:conditional-results}
\end{figure}

Figure \ref{fig:conditional-results} shows some example of synthetic food image resulted from conditional StyleGAN3. However, the generated images do not look realistic and contain obvious visual artifacts. 

\begin{figure}[h!]
  \centering
  \includegraphics[width=0.7\columnwidth]
  {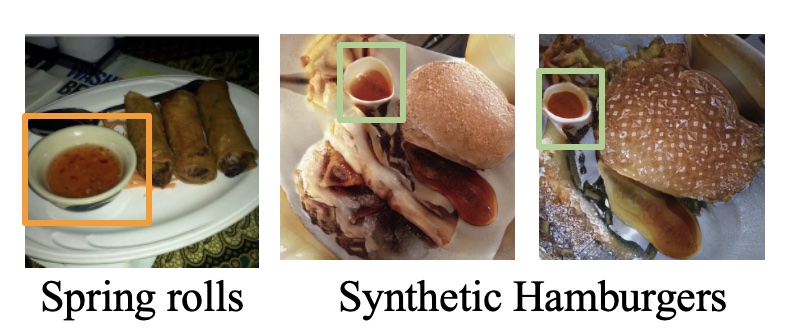}
  \caption{\textbf{An Illustration of Inter-class Feature Entanglement Issue in Synthetic Hamburger Images}}
  \label{fig:inter-class-issue}
\end{figure}

The most obvious artifact we notice in those generated images is the inter-class feature entanglement. For example, in Figure \ref{fig:inter-class-issue}, the small disc-shape feature appears in the synthetic hamburger images, but it should only appear in the spring roll food images. This issue can be resolved when we only train one class at a time.

\subsection{Results on Low-resolution Single-class Food Datasets}
\label{subsec: low-single}
\begin{figure}[!hb]
  \centering
  \includegraphics[width=0.9\columnwidth]
  {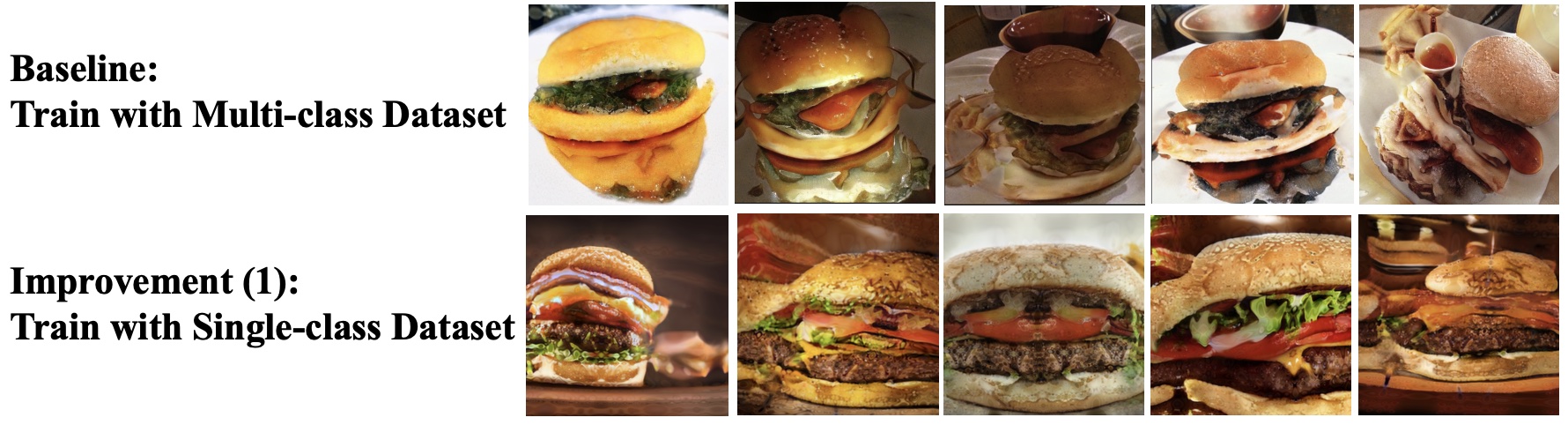}
  \caption{\textbf{Sample Synthetic Hamburger images from Baseline and Improved Methods}}
  \label{fig:compare-results}
\end{figure}

Figure \ref{fig:compare-results} shows the comparison results of synthetic hamburger images between multi-class and single-class trained on StyleGAN3. Without inter-class feature entanglement, our generator only captures in-class features and the synthetic hamburger image results are much more realistic compared to the baseline of training with multiple food classes. 

\begin{figure}[t]
  \centering
  \includegraphics[width=0.7\columnwidth]
  {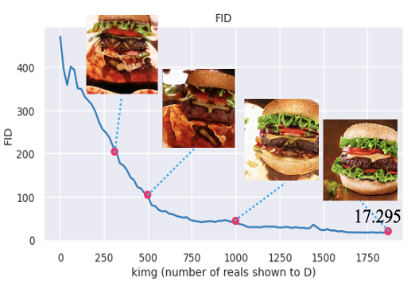}
  \caption{\textbf{Evaluation of Synthetic Hamburger Images on StyleGAN3}}
  \label{fig:fid-eval}
\end{figure}

Similar to the multi-class training, we train the network for about 2,000 iterations for the network to converge at 17.295. With the trained StyleGAN3 model on hamburger image samples, we use it as our pretrained model for our next phase of any-resolution training. Results are shown in Figure \ref{fig:fid-eval}. Although the FID score is similar to training on multi-class food images, the visual artifacts are significantly reduced. 

\subsection{Results on Any-resolution Single-class Food Datasets}
\label{subsec: any-single}
\begin{figure}[t]
  \centering
  \includegraphics[width=0.9\columnwidth]
  {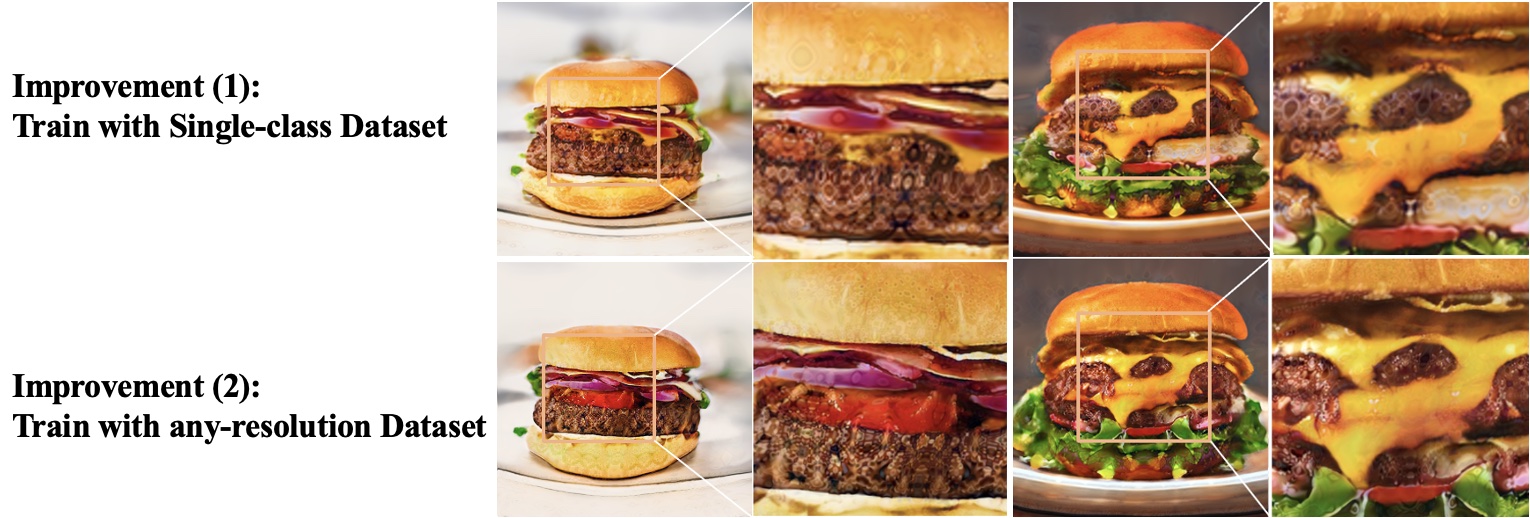}
  \caption{\textbf{Comparison between two improved methods on StyleGAN3}}
  \label{fig:two-method-compare}
\end{figure}

To avoid image warping and loss of high-resolution details during image normalization using StyleGAN3’s fixed resolution, we train square-shaped image patches cropped from any-resolution datasets.  Following the two-phase training of any-resolution dataset, we train the StyleGAN3 model with $256 \times 256$ low-resolution hamburger images from Food-101 for the first stage of StyleGAN3 pretraining and our HR dataset for second stage image-patches training. Figure \ref{fig:two-method-compare} shows the comparison of the synthetic hamburger images between the first improvement on StyleGAN3 with single-hamburger class training and the second improvement with our any-resolution training. The visual quality of hamburgers is further improved, and details are better preserved.

\vspace{-0.3cm}
\subsection{Quantitative Results}
\label{subsec: objective}
As shown in Table \ref{table:fid-eval}, we calculate the standard FID metric for the first and second improvements made to the StyleGAN3 training strategy. The standard FID metrics between training using single-class and image-patches are similar, despite the obvious visual improvement from image-patches training. This is because the standard FID metric assumes all the training images are of $256 \times 256$ resolution, which ignores the fine-grain details in the training dataset. Thus, the standard FID scores is not suitable for evaluating our image-patches training results. Instead, we adopt the patch-FID (pFID) metric, which extracts 50K image patches cropped from our Any-resolution dataset at various scales and locations. To avoid downsampling the training images, it computes the FID score on the generated patches and real patches with corresponding scales and locations. The pFID score in Table \ref{table:fid-eval} confirms our observation that with image-patches training, the food images contain details and are visually more realistic.

\begin{table}[!h]
\caption{FID and patch-FID Metric Evaluation on Two methods at 256 Image Resolution}
\label{tab:fonts}
\begin{center}       
\begin{tabular}{p{0.55\columnwidth} p{0.15\columnwidth} p{0.15\columnwidth}} 
\hline
\textbf{Improvement Methods} & \textbf{FID} & \textbf{pFID}\\ \hline
Train with Single-class Dataset & 17.871 & 90.113\\ 
Train with any-resolution Dataset & 17.723 & \textbf{30.863}\\ \hline
\label{table:fid-eval}
\end{tabular}
\end{center}
\end{table} 

\vspace{-1.5cm}
\subsection{Subjective Study}
\label{subsec: subjective}


We conduct a subjective study to assess the perceptual realism of synthetically generated food images to qualitatively evaluate our conditional synthetic food image generation model. This subjective measure is an important complement to the Frechet Inception Distance (FID). The synthetic food image should look realistic so that it can be used for downstream tasks such as food image classification as training examples. In the survey, 82 adult participants were asked to evaluate 88 food images which contain 51 synthetic food images and 37 real food images. The synthetic images are evenly distributed among three classes –- hamburger, pizza, and spring roll. For real food images, we select 12 images of hamburgers, 12 images of pizzas, and 13 images of spring rolls. Participants were asked to select images that looked real to them (\textit{i.e.}, did not look synthetic) and were asked to look at each image for no more than 3 seconds. We also set a scoring system to evaluate our model performance –- 51 is the full score since they are 51 synthetic images, and participants received one point for selecting the synthetic image.

On average, participants scored 33.02 out of 51, which means that they mistook 64.75\% of the generated synthetic food images as real images. Every synthetic food image has at least twenty-five participants who thought it is real. Among the 17 synthetic pizza images, on average, participants selected 45.65\% of them as real images. Among the 17 synthetic hamburger images, on average, participants selected 38.29\% of them as real images. Among the 17 synthetic spring roll images, on average, participants selected 52.17\% of them as real images. From our survey results, more than half of our generated synthetic images are realistic enough to make participants select them as real images. We can conclude from the subjective study that our proposed methods can effectively learn realistic features from the real samples and may be good enough to be used as representative training examples for downstream tasks, such as food image recognition when there is a lack of training images. 


\subsection{Impact of Using Synthetic Images on Food Classification}
\label{subsec: downstream}
Most deep learning-based methods require a large number of training data which can be challenging for many applications. Synthetic images that closely resemble real ones could be a potential resolution to address this problem. In order to assess the effect of using synthetic images as part of training data, we design experiments to explore the impact of synthetic images as data augmentation for the food image classification task. 


Our experiment aims to classify food images from three different classes (hamburger, pizza, and spring roll). The images we used for training are either randomly picked from the Food-101 dataset (LR dataset) or high-resolution food images from the Any-resolution dataset. We consider 3 experimental setups as described below.
\begin{enumerate}[noitemsep,topsep=0pt]
    \item We train the ResNet-50 with only 200 real food images (100 from LR and 100 from the Any-resolution dataset).
    \item We train the same model with 200 real images from the first experiment and an extra 200 of our generated synthetic images.
    \item We train the same model with the same 200 real images from experiment one and an extra 200 real images (100 from LR and 100 from the Any-resolution dataset). (Upper bound)
\end{enumerate}
All the three experiments use the same testing set containing 100 images (50 from LR and 50 from the Any-resolution dataset). We apply ResNet-50 as the backbone and keep the training settings the same with a batch size of 64, and training epoch around 100 for all three experiments to ensure a fair comparison. 

\begin{figure}[t]
  \centering
  \includegraphics[width=0.8\columnwidth]
  {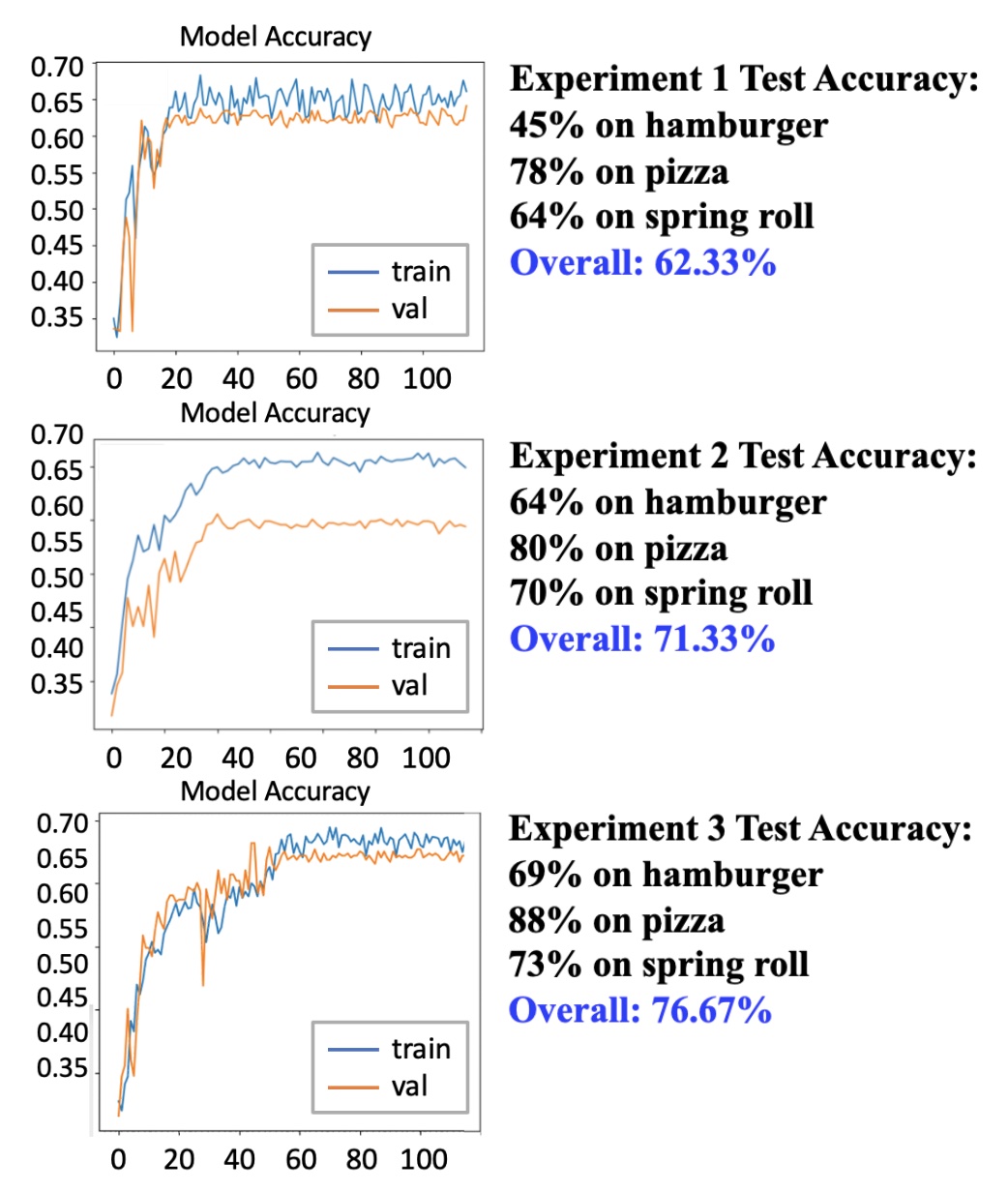}
  \caption{\textbf{ResNet-50 Training (Blue) and Validation (Orange) Accuracy for Three Different Experiments}}
  \label{fig:resnet-result}
\end{figure}


Figure \ref{fig:resnet-result} shows the comparison of the best training, validation, and testing accuracy results from three different experiments. As expected from those plotted figures, Experiment 1 has the overall lowest accuracy, \textit{i.e.} 62.33\%, since it uses the least amount of training data. Experiment 2 achieves 71.33\% accuracy and greatly improves the model by almost 10\% by training with the additional synthetic images. Finally, Experiment 3 (upper bound) has the highest testing accuracy, 76.67\%. As observed from the plotted figure of Experiment 2, the validation accuracy is lower than the training accuracy, which is caused by different data distribution between training and testing datasets since our generated synthetic images still contain unnatural artifacts. Nonetheless, our preliminary experiments demonstrate that using synthetic images to augment datasets is effective in improving the model's performance on the food image classification task.  Furthermore, compared to pizza, hamburgers and spring rolls have lower accuracy due to their more complex and dynamic features, which is still difficult for our food image generation model to produce good feature representation. This is consistent with our visual observation that the synthetic images of hamburgers and spring rolls are less realistic than pizzas. 
Overall, we show that the generated synthetic images are realistic enough to be used as training samples when real data is scarce and can greatly improve the performance of a deep learning model (in our case food classification).

\section{Conclusion}
In this paper, we propose an improved conditional synthetic food image generation based on the StyleGAN3 baseline method. The first improvement uses single-class training instead of multi-class to avoid the inter-class feature entanglement. Next, we leverage square-shaped image patches training to retain high-resolution details in our generated images as opposed to a fixed resolution input. With our improved methods, our synthetic food image generation results are more realistic and contain more details compared to the baseline method. In addition to the quantitative evaluation of our proposed method, we conduct a subject study to qualitatively assess the perceptual realism of generated synthetic images. On average, participants mistaken 64.75\% of the generated synthetic food images as real images. To show the impact of synthetic images for downstream tasks, we conducted a set of experiments where synthetic images were used to augment training data for the food image classification task using a ResNet-50 model. Results show significant improvement in classification accuracy. Our future work will focus on developing a multi-label training strategy to generate multiple food classes in a single image and apply it to other vision tasks such as food image localization and volume estimation.

\vspace{-0.4cm}
\bibliographystyle{IEEEbib}
{\small\bibliography{ref}}

\vspace{-0.5cm}
\begin{biography}
Wenjin Fu received her B.S. degree with Cum Laude honor from The Ohio State University in December 2022. She is currently studying in Electrical and Computer Engineer MS program at Carnegie Mellon University. Her research interests are in Artificial Intelligence applications specifically in computer vision, software development, and autonomy for mobile.
\newline

Yue Han received his B.S degree with distinction from Purdue University in 2019. He is currently pursuing a Ph.D. degree at Purdue University and working as a research assistant in the Video and Image Processing (VIPER) Laboratory at Purdue University. His research interests include image processing, computer vision, and deep learning.
\newline

Jiangpeng He received his Ph.D. degree in Electrical and Electronic Engineering from Purdue University  in August 2022. He is currently a postdoc research assistant at the School of Electrical and Computer Engineering, Purdue University, West Lafayette, IN, USA. His research interests include image processing, computer vision, image-based dietary assessment and deep learning.
\newline

Sriram Baireddy is a Ph.D. candidate in Electrical Engineering at Purdue University. He earned his B.S. and M.S. degrees in Electrical Engineering at Purdue in 2018 and 2021, respectively, with minors in economics, math, and physics. He currently investigates the application of machine learning techniques to signals, images, and videos for forensic and agricultural research.
\newline

Mridul Gupta received his B.Tech degree in Civil Engineering from Indian Institute of Technology Roorkee in 2017. He is currently a Ph.D. candidate in Video and Image Processing Lab (VIPER) advised by Prof. Edward J. Delp at Purdue University. His research  focuses on applying deep learning and machine learning tools to problems in computer vision and image processing. \newline

\textbf{Fengqing Zhu} is an Associate Professor of Electrical and Computer Engineering at Purdue University, West Lafayette, Indiana. She received the B.S.E.E. (with highest distinction), M.S. and Ph.D. degrees in Electrical and Computer Engineering from Purdue University. She is the recipient of an NSF CISE Research Initiation Initiative (CRII) award in 2017, a Google Faculty Research Award in 2019, and an ESI and trainee poster award for the NIH Precision Nutrition workshop in 2021. She is a senior member of the IEEE.

\end{biography}

\end{document}